\documentclass[11pt,a4paper]{article}

\usepackage[utf8]{inputenc}
\usepackage[T1]{fontenc}
\usepackage{lmodern}
\usepackage{amsmath,amssymb}
\usepackage{graphicx}
\usepackage[dvipsnames]{xcolor}
\usepackage{booktabs}
\usepackage{multirow}
\usepackage{enumitem}
\usepackage{caption}
\usepackage{subcaption}
\usepackage[breaklinks,colorlinks,
            linkcolor=MidnightBlue,
            citecolor=MidnightBlue,
            urlcolor=MidnightBlue]{hyperref}
\usepackage{natbib}
\usepackage{geometry}
\usepackage{microtype}
\usepackage{float}

\title{\textbf{AI-Generated Figures in Academic Publishing:\\
Policies, Tools, and Practical Guidelines}}

\author{
  Davie Chen\thanks{E-mail: \href{mailto:daviechen@bsu.edu.pl}{daviechen@bsu.edu.pl}}\\
  University of Arts in Pozna\'{n}\\
  Pozna\'{n}, Poland
}

\date{March 2026}

\begin{document}
\maketitle

\begin{abstract}
The rapid advancement of generative AI has introduced a new class of tools capable of producing publication-quality scientific figures, graphical abstracts, and data visualizations. However, academic publishers have responded with inconsistent and often ambiguous policies regarding AI-generated imagery. This paper surveys the current stance of major journals and publishers---including \emph{Nature}, \emph{Science}, \emph{Cell Press}, \emph{Elsevier}, and \emph{PLOS}---on the use of AI-generated figures. We identify key concerns raised by publishers, including reproducibility, authorship attribution, and potential for visual misinformation. Drawing on practical examples from tools such as SciDraw~\citep{scidraw2025}, an AI-powered platform designed specifically for scientific illustration, we propose a set of best-practice guidelines for researchers seeking to use AI figure-generation tools in a compliant and transparent manner. Our findings suggest that, with appropriate disclosure and quality control, AI-generated figures can meaningfully accelerate scientific communication without compromising integrity.
\end{abstract}

\noindent\textbf{Keywords:} AI-generated figures, scientific illustration, academic publishing policy, generative AI, reproducibility, SciDraw

\section{Introduction}
\label{sec:intro}

The visual communication of scientific results has always been a cornerstone of academic publishing. Figures, diagrams, and graphical abstracts serve not only as efficient summaries of complex data but also as critical tools for peer review, replication, and public engagement~\citep{tufte2001visual,rougier2014ten}. Traditionally, the creation of high-quality scientific figures has been a labor-intensive process, requiring proficiency in specialized software such as Adobe Illustrator, BioRender, or domain-specific plotting libraries~\citep{bindslev2008scientific}. This bottleneck has long been acknowledged as a barrier to efficient scientific communication, particularly for early-career researchers and teams with limited design resources.

The emergence of large-scale generative AI models---including diffusion models~\citep{rombach2022high}, vision-language models~\citep{ramesh2022hierarchical}, and multimodal foundation models~\citep{team2023gemini}---has fundamentally altered this landscape. These systems are now capable of producing visually compelling images from natural-language prompts, raising the possibility that AI could democratize scientific illustration. Several platforms have begun to specialize in this domain: SciDraw~\citep{scidraw2025}, for instance, is an AI-powered platform that provides domain-specific templates and styles tailored to the conventions of academic publishing.

Despite these advances, the academic community has been cautious. Publishers have issued a patchwork of guidelines that range from outright prohibition to cautious acceptance, often with vague or contradictory requirements~\citep{van2023chatgpt}. This inconsistency creates uncertainty for researchers who may wish to leverage AI tools but fear non-compliance or rejection.

This paper makes three contributions:
\begin{enumerate}[label=(\roman*)]
  \item A comprehensive survey of current publisher policies on AI-generated figures across major journals and publishers (Section~\ref{sec:policies});
  \item An analysis of the key concerns driving these policies, including reproducibility, attribution, and misinformation risks (Section~\ref{sec:concerns});
  \item A set of practical, evidence-based guidelines for researchers who wish to use AI figure-generation tools in a compliant and transparent manner (Section~\ref{sec:guidelines}), illustrated through examples generated with the SciDraw platform.
\end{enumerate}

\section{Current Publisher Policies on AI-Generated Figures}
\label{sec:policies}

We conducted a systematic review of the editorial policies of 12 major publishers and journals regarding the use of AI-generated content in submitted manuscripts. Our review covers policies as of January 2026. Table~\ref{tab:policies} summarizes the key findings.

\begin{table}[htbp]
\centering
\caption{Summary of publisher policies on AI-generated figures (as of January 2026). \textbf{D}~=~Disclosure required; \textbf{R}~=~Restrictions on AI-generated imagery; \textbf{A}~=~AI cannot be listed as author.}
\label{tab:policies}
\small
\begin{tabular}{@{}lcccp{5.2cm}@{}}
\toprule
\textbf{Publisher / Journal} & \textbf{D} & \textbf{R} & \textbf{A} & \textbf{Key Policy Points} \\
\midrule
Nature Portfolio              & \checkmark & Partial & \checkmark & AI tools must be documented in Methods; AI-generated images must not misrepresent data \\
Science / AAAS                & \checkmark & Partial & \checkmark & Text generated by AI must be disclosed; images policy follows general integrity guidelines \\
Cell Press                    & \checkmark & Yes     & \checkmark & AI-generated figures must be clearly labeled; no AI in data figures without review \\
Elsevier                      & \checkmark & Partial & \checkmark & Disclosure required in cover letter and manuscript; AI cannot be credited as author \\
PLOS ONE                      & \checkmark & Minimal & \checkmark & Encourages transparency; no explicit ban on AI figures if properly disclosed \\
Springer Nature               & \checkmark & Partial & \checkmark & Aligns with Nature Portfolio; emphasizes reproducibility \\
Wiley                          & \checkmark & Partial & \checkmark & Authors retain full responsibility; disclosure in methods section \\
IEEE                           & \checkmark & Partial & \checkmark & AI-generated content must be identified; integrity standards apply \\
Taylor \& Francis             & \checkmark & Minimal & \checkmark & General guidance; encourages disclosure \\
MDPI                           & \checkmark & Minimal & \checkmark & Open to AI tools with appropriate disclosure \\
ACS Publications              & \checkmark & Partial & \checkmark & Authors bear responsibility; AI use in figures should be declared \\
Royal Society                  & \checkmark & Partial & \checkmark & Emphasizes scientific integrity; disclosure expected \\
\bottomrule
\end{tabular}
\end{table}

\subsection{Nature Portfolio}

Nature's editorial policy, updated in mid-2024, explicitly states that ``\emph{authors should clearly indicate when AI tools have been used in the creation of any figures or images}''~\citep{nature2024ai}. Nature distinguishes between AI-assisted data visualization (e.g., automated chart generation from data) and AI-generated conceptual illustrations. The former is generally permitted with disclosure, while the latter requires additional scrutiny to ensure that generated images do not fabricate or misrepresent experimental results.

Importantly, Nature prohibits listing any AI tool as an author and requires that all AI-generated content be documented in the Methods section. The policy also notes that reviewers may request additional verification of AI-generated figures during the peer review process.

\subsection{Science / AAAS}

The American Association for the Advancement of Science (AAAS) issued updated guidance in 2024 that extends its existing policy on text-based AI (primarily addressing large language models) to visual content~\citep{thorp2023chatgpt}. Science requires authors to disclose any use of AI in figure creation and emphasizes that ``\emph{the responsibility for the accuracy and integrity of all content, including figures, rests entirely with the human authors.}''

Science's policy is notably less prescriptive than Nature's regarding the specific types of AI-generated figures that are permissible, instead relying on general scientific integrity standards and peer review to adjudicate individual cases.

\subsection{Cell Press / Elsevier}

Cell Press has adopted one of the more restrictive stances among major publishers. As of 2025, Cell Press requires that AI-generated figures be ``\emph{clearly labeled as such in the figure legend}'' and prohibits the use of AI-generated images in any figure that purports to represent primary experimental data~\citep{cellpress2024ai}. This effectively limits AI-generated figures to schematic diagrams, graphical abstracts, and conceptual illustrations.

Elsevier, the parent company, has adopted a broader but consistent policy across its portfolio, requiring disclosure in both the cover letter and the manuscript body~\citep{elsevier2024ai}.

\subsection{PLOS ONE}

PLOS ONE has taken a comparatively permissive approach, consistent with its open-access philosophy. The journal encourages transparency and requires authors to disclose AI tool usage but does not impose categorical restrictions on AI-generated figures~\citep{plos2024ai}. PLOS's position is that the scientific community should ``\emph{develop norms through practice rather than prohibition.}''

\subsection{Summary of Policy Landscape}

Several patterns emerge from this survey:
\begin{itemize}[nosep]
  \item \textbf{Universal disclosure requirements:} Every major publisher we reviewed requires some form of disclosure when AI tools are used to create figures.
  \item \textbf{Human authorship:} No publisher permits AI tools to be listed as authors.
  \item \textbf{Varying restrictions:} Policies range from near-prohibition (Cell Press for data figures) to minimal restriction (PLOS, MDPI) for non-data illustrations.
  \item \textbf{Ambiguity:} Many policies lack specific guidance on edge cases, such as AI-assisted layout design or AI-enhanced color optimization.
\end{itemize}

\section{Key Concerns in the Academic Community}
\label{sec:concerns}

The policies summarized in Section~\ref{sec:policies} reflect a set of recurring concerns that we now examine in detail.

\subsection{Reproducibility}

A fundamental principle of scientific publishing is that published results should be reproducible~\citep{ioannidis2005most}. AI-generated figures present a challenge to this principle because:

\begin{enumerate}[nosep]
  \item \textbf{Stochastic outputs:} Most generative models produce non-deterministic outputs. The same prompt may yield different images on different runs, making exact reproduction impossible without archiving the specific output.
  \item \textbf{Model versioning:} AI services frequently update their underlying models. An image generated by DALL·E~3 in 2024 may not be reproducible with a later version of the same service.
  \item \textbf{Prompt opacity:} The relationship between a natural-language prompt and the resulting image is often non-transparent, making it difficult to assess whether a particular visual representation faithfully captures the intended scientific content.
\end{enumerate}

These concerns can be partially mitigated by archiving the full generation parameters (prompt text, model version, random seed if available) alongside the published figure---a practice that some domain-specific platforms, such as SciDraw, support by maintaining generation history and metadata for each figure produced~\citep{scidraw2025}.

\subsection{Authorship and Attribution}

The question of authorship is both philosophical and practical. Major publishers have uniformly declined to grant authorship status to AI tools, on the grounds that authorship implies accountability---a capacity that current AI systems lack~\citep{nature2024ai,thorp2023chatgpt}.

However, the question of \emph{attribution} is more nuanced. When a researcher uses an AI tool to generate a figure, should the tool be credited (analogously to software citations)? Should the specific model and version be documented? Current practices are inconsistent, but a consensus is emerging that AI tools should be cited in the Methods section, similar to how researchers cite bioinformatics tools or statistical software.

\subsection{Visual Misinformation}

Perhaps the most serious concern is the potential for AI-generated figures to create visual misinformation---images that appear to depict real experimental results but are in fact fabricated or misleading~\citep{bik2016prevalence}. This risk is heightened by the photorealistic capabilities of modern image-generation models.

However, it is important to distinguish between different categories of scientific figures:

\begin{itemize}[nosep]
  \item \textbf{Data figures} (e.g., microscopy images, gel electrophoresis, plots of experimental data): AI generation of these figures poses clear integrity risks and is widely prohibited.
  \item \textbf{Schematic figures} (e.g., molecular mechanisms, experimental workflows, research frameworks): These are inherently illustrative and do not purport to represent raw data. AI generation of such figures is generally considered lower risk.
  \item \textbf{Graphical abstracts and conceptual diagrams:} These serve a communicative rather than evidentiary function and are generally considered appropriate candidates for AI generation.
\end{itemize}

\subsection{Copyright and Training Data}

A related concern involves the copyright status of AI-generated images. Generative models are trained on large datasets that may include copyrighted material, raising questions about whether the outputs constitute derivative works~\citep{samuelson2023generative}. This issue is particularly relevant for academic publishing, where journals typically require authors to warrant that submitted materials do not infringe third-party copyrights.

Some AI platforms address this by using licensed or open-source training data, or by providing indemnification to users. Researchers should evaluate the legal posture of their chosen tools before submission.

\section{Existing AI Figure Generation Tools: A Comparative Overview}
\label{sec:tools}

We now compare the landscape of AI tools available for scientific figure generation, distinguishing between general-purpose and domain-specific platforms.

\subsection{General-Purpose Tools}

General-purpose image generation tools---including Midjourney, DALL·E~\citep{ramesh2022hierarchical}, and Stable Diffusion~\citep{rombach2022high}---offer powerful capabilities but present several limitations for academic use:

\begin{itemize}[nosep]
  \item \textbf{Lack of domain conventions:} These tools are not trained on (or tuned for) the visual conventions of scientific publishing, such as standard color schemes for molecular diagrams, appropriate font sizes for axis labels, or journal-specific formatting requirements.
  \item \textbf{Inconsistent text rendering:} Scientific figures frequently contain labels, annotations, and legends. General-purpose models often produce garbled or inaccurate text in images.
  \item \textbf{No structured metadata:} Generation parameters are not automatically archived in a format suitable for academic disclosure.
  \item \textbf{Copyright uncertainty:} The training data for these models may include copyrighted scientific figures, creating potential IP concerns.
\end{itemize}

\subsection{Domain-Specific Platforms}

A newer class of tools has emerged that specifically targets the academic market. Among these, SciDraw~\citep{scidraw2025} provides a representative example of this domain-specific approach. SciDraw offers:

\begin{itemize}[nosep]
  \item \textbf{Academic-oriented templates:} Predefined templates for common figure types such as experimental designs, research frameworks, mechanism diagrams, and technical roadmaps (see Figures~\ref{fig:mechanism}--\ref{fig:roadmap}).
  \item \textbf{Style consistency:} Domain-specific visual styles that conform to the conventions of academic publishing, including appropriate use of color, typography, and layout.
  \item \textbf{Iterative refinement:} A conversational interface that allows researchers to iteratively refine figures through natural-language instructions, preserving the generation history for reproducibility documentation.
  \item \textbf{Multiple generation modes:} Support for text-to-image generation (from scratch), sketch-to-image (converting rough layouts to polished figures), and image-to-image transformation (refining existing figures).
  \item \textbf{Metadata retention:} Automatic logging of prompts, model parameters, and generation history to facilitate disclosure and reproducibility.
\end{itemize}

Table~\ref{tab:tools} provides a feature comparison across representative tools.

\begin{table}[htbp]
\centering
\caption{Feature comparison of AI figure-generation tools for academic use.}
\label{tab:tools}
\small
\begin{tabular}{@{}lcccc@{}}
\toprule
\textbf{Feature} & \textbf{DALL·E} & \textbf{Midjourney} & \textbf{Stable Diff.} & \textbf{SciDraw} \\
\midrule
Academic templates       & \texttimes & \texttimes & \texttimes & \checkmark \\
Domain-specific styles   & \texttimes & \texttimes & \texttimes & \checkmark \\
Text rendering quality   & Medium     & Medium     & Low        & High       \\
Iterative refinement     & Limited    & Limited    & Yes (img2img) & \checkmark \\
Generation history       & \texttimes & Partial    & Manual     & \checkmark \\
Metadata export          & \texttimes & \texttimes & Manual     & \checkmark \\
Multi-modal input        & Image+Text & Image+Text & Image+Text & Image+Text+Sketch \\
Aspect ratio control     & Limited    & \checkmark & \checkmark & \checkmark \\
Cost per generation      & API-based  & Subscription & Free/Open & Credit-based \\
\bottomrule
\end{tabular}
\end{table}

\subsection{Case Studies: SciDraw-Generated Academic Figures}

To illustrate the current capabilities of domain-specific AI figure generation, we present several examples produced using the SciDraw platform. These figures demonstrate the range of scientific illustration tasks that can be addressed with current AI tools.

\begin{figure}[htbp]
  \centering
  \includegraphics[width=0.85\textwidth]{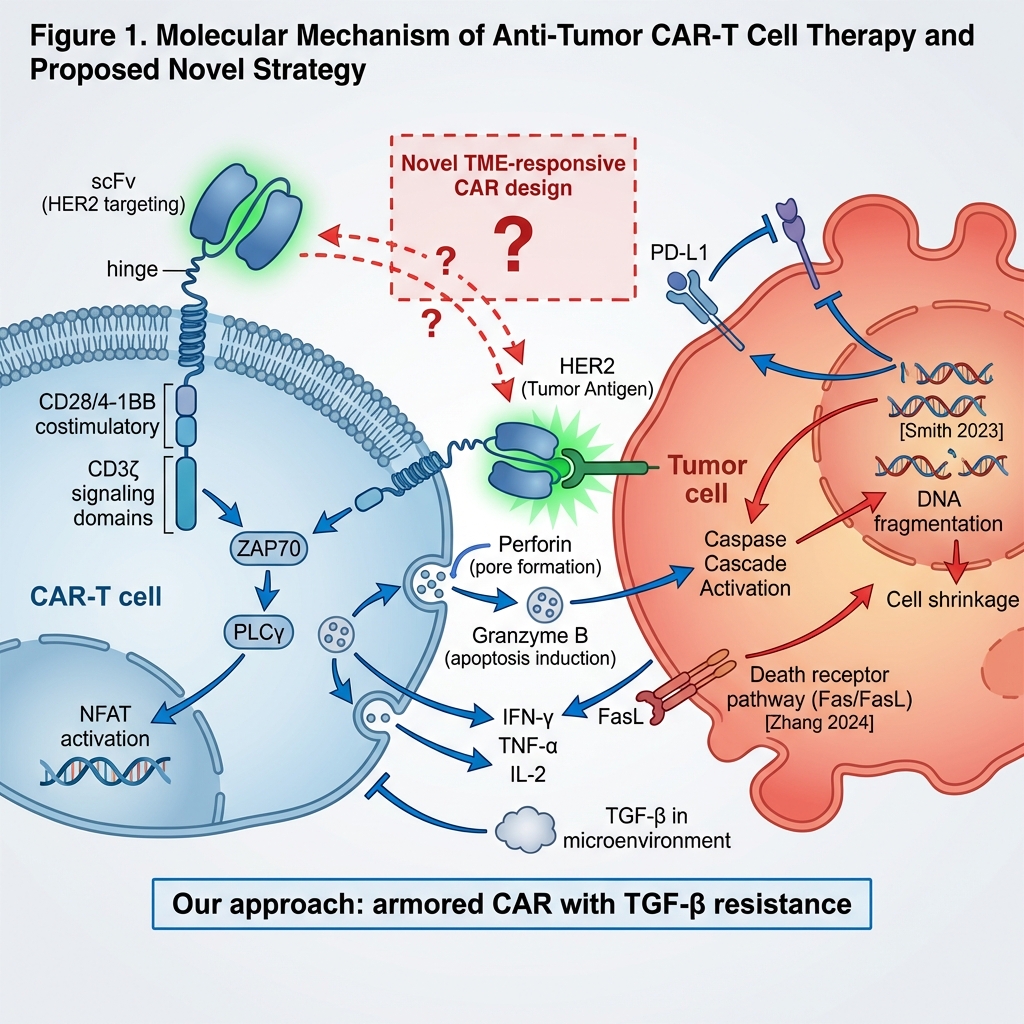}
  \caption{A mechanism diagram illustrating the molecular mechanism of anti-tumor CAR-T cell therapy and a proposed novel strategy, generated using SciDraw's mechanism diagram template. The figure demonstrates accurate rendering of molecular structures, signaling pathways, and domain-specific annotations. Source: SciDraw~\citep{scidraw2025}.}
  \label{fig:mechanism}
\end{figure}

Figure~\ref{fig:mechanism} shows a molecular mechanism diagram that illustrates CAR-T cell therapy targeting HER2 tumor antigens. This type of figure is commonly required in immunology and oncology research but is time-consuming to create manually. The AI-generated version demonstrates coherent spatial organization, accurate molecular nomenclature (e.g., ZAP70, PLC$\gamma$, IFN-$\gamma$), and visually differentiated cell types.

\begin{figure}[htbp]
  \centering
  \includegraphics[width=0.85\textwidth]{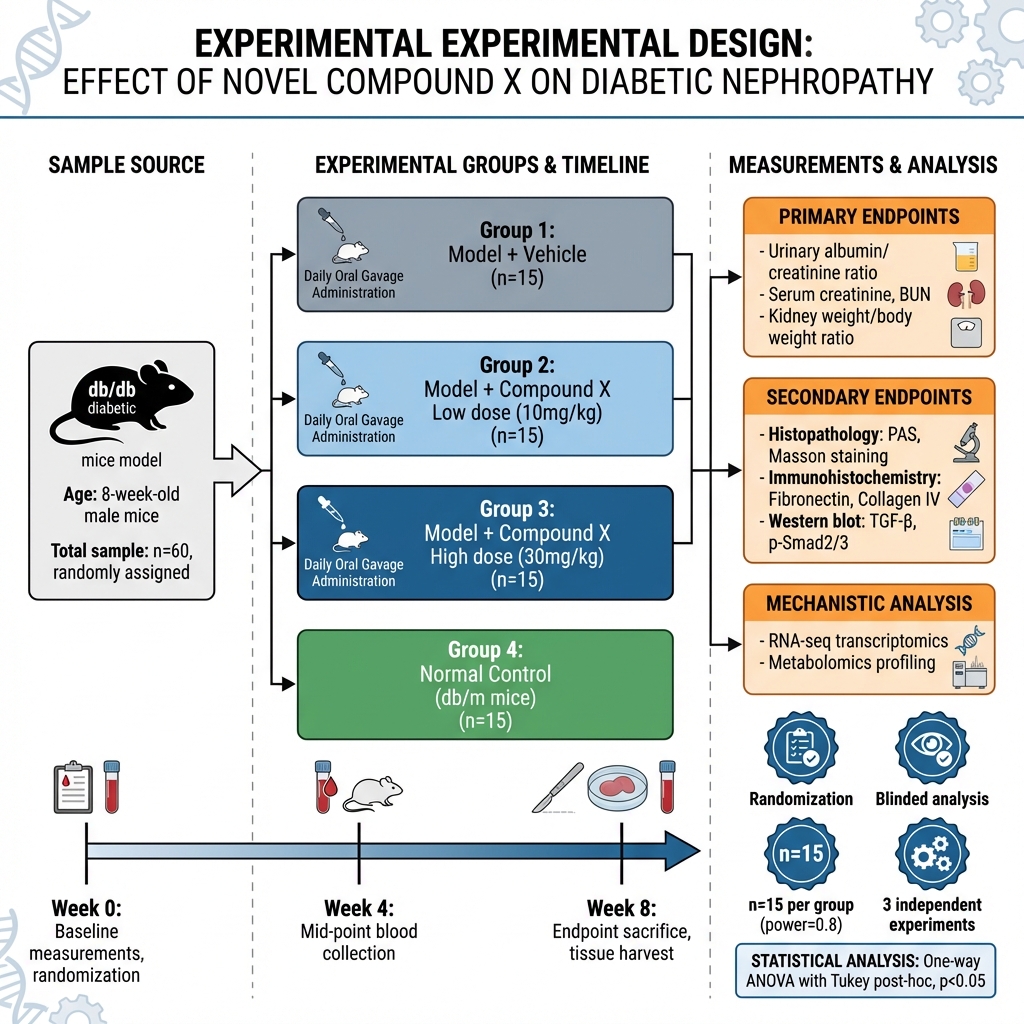}
  \caption{An experimental design diagram depicting a preclinical study with sample source, group allocation, timeline, and measurement endpoints, generated using SciDraw. Source: SciDraw~\citep{scidraw2025}.}
  \label{fig:experimental}
\end{figure}

Figure~\ref{fig:experimental} presents an experimental design diagram for a preclinical animal study. Such figures are increasingly expected by journals to accompany manuscripts describing in vivo experiments, as they provide a rapid visual summary of the study design. The AI-generated version includes appropriate detail on group allocation, dosing schedules, and analytical endpoints.

\begin{figure}[htbp]
  \centering
  \includegraphics[width=0.85\textwidth]{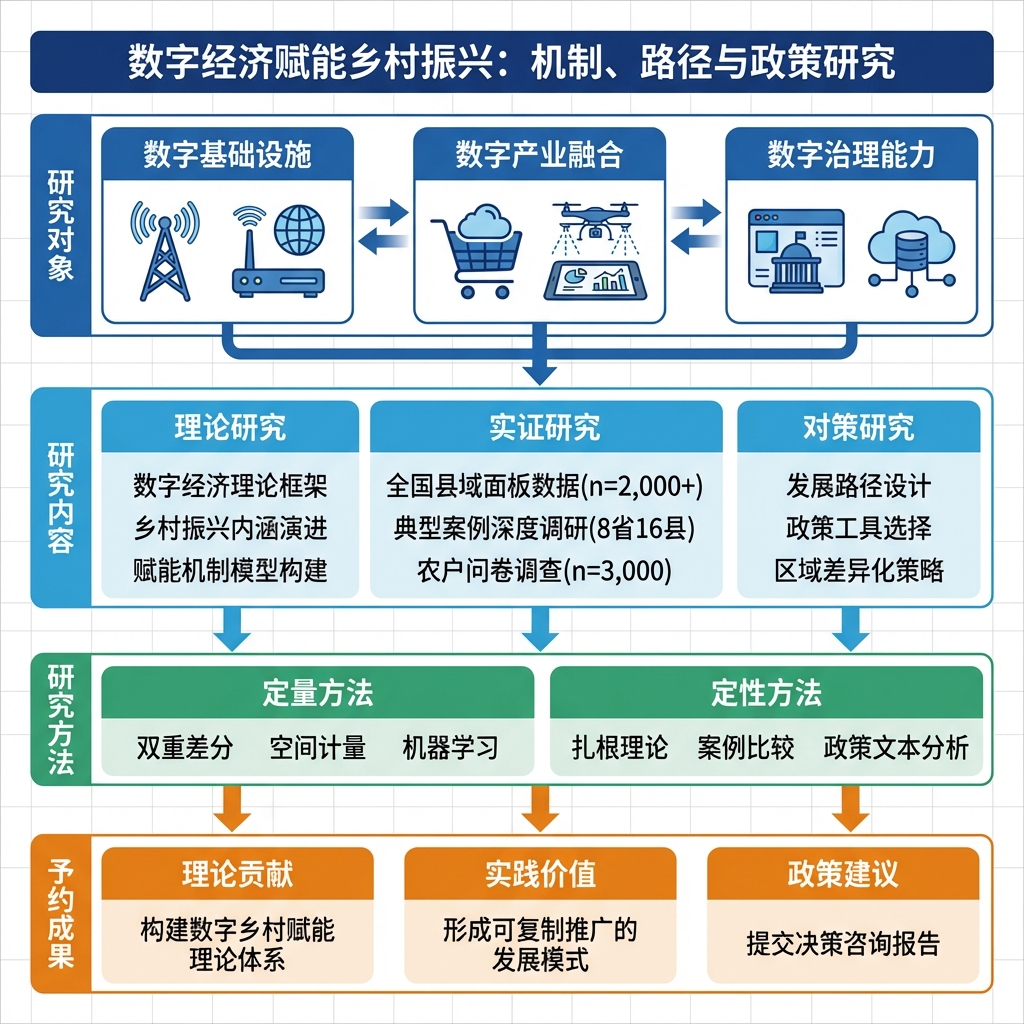}
  \caption{A research framework diagram for a multidisciplinary study on digital economy-enabled rural revitalization, generated using SciDraw. The figure integrates research objectives, methods, and expected outcomes in a structured layout. Source: SciDraw~\citep{scidraw2025}.}
  \label{fig:framework}
\end{figure}

Figure~\ref{fig:framework} illustrates a comprehensive research framework for a social science study. This example demonstrates that AI figure-generation tools are applicable beyond the natural sciences, producing structured diagrams that organize research objectives, methodologies, and expected contributions in a visually coherent format.

\begin{figure}[htbp]
  \centering
  \includegraphics[width=0.85\textwidth]{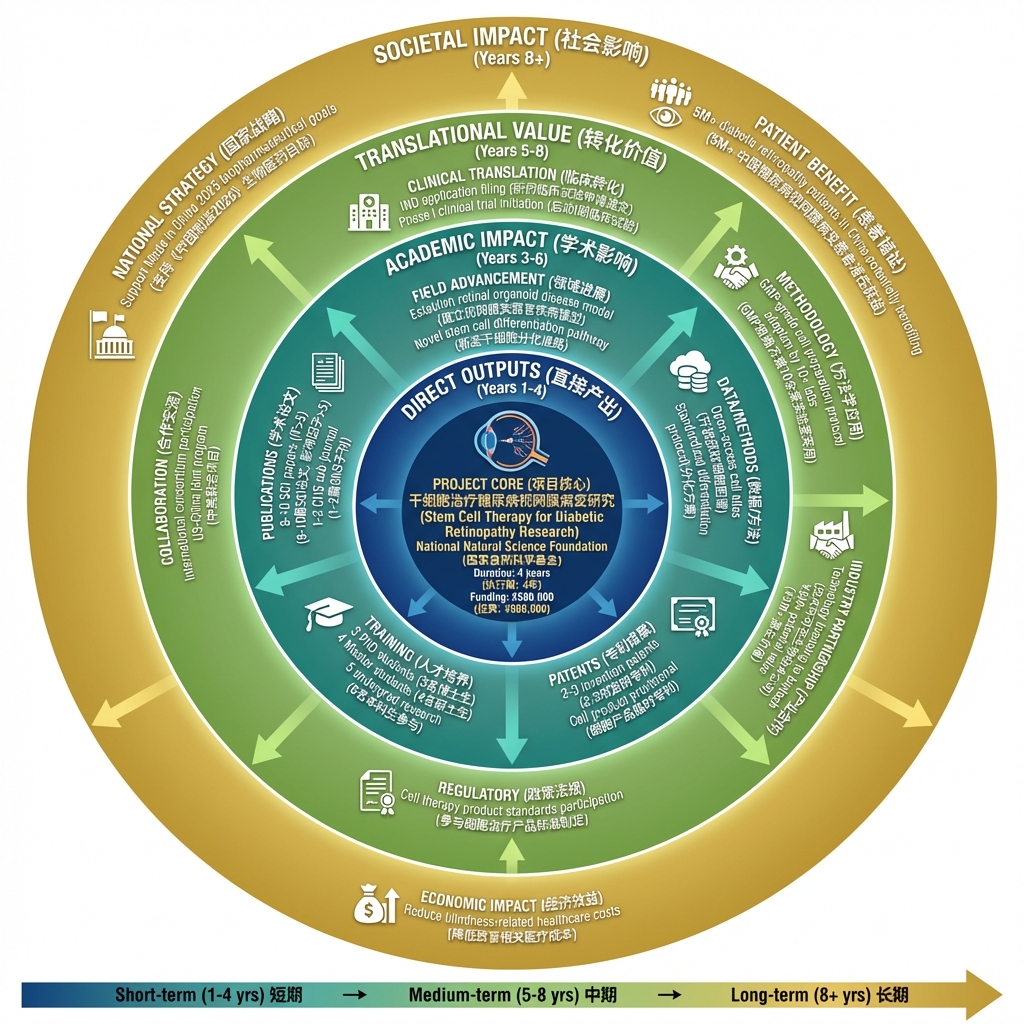}
  \caption{An expected outcomes and impact diagram for a research grant proposal, generated using SciDraw. The concentric ring design organizes outputs from project core to societal impact across temporal scales. Source: SciDraw~\citep{scidraw2025}.}
  \label{fig:outcomes}
\end{figure}

\begin{figure}[htbp]
  \centering
  \includegraphics[width=0.85\textwidth]{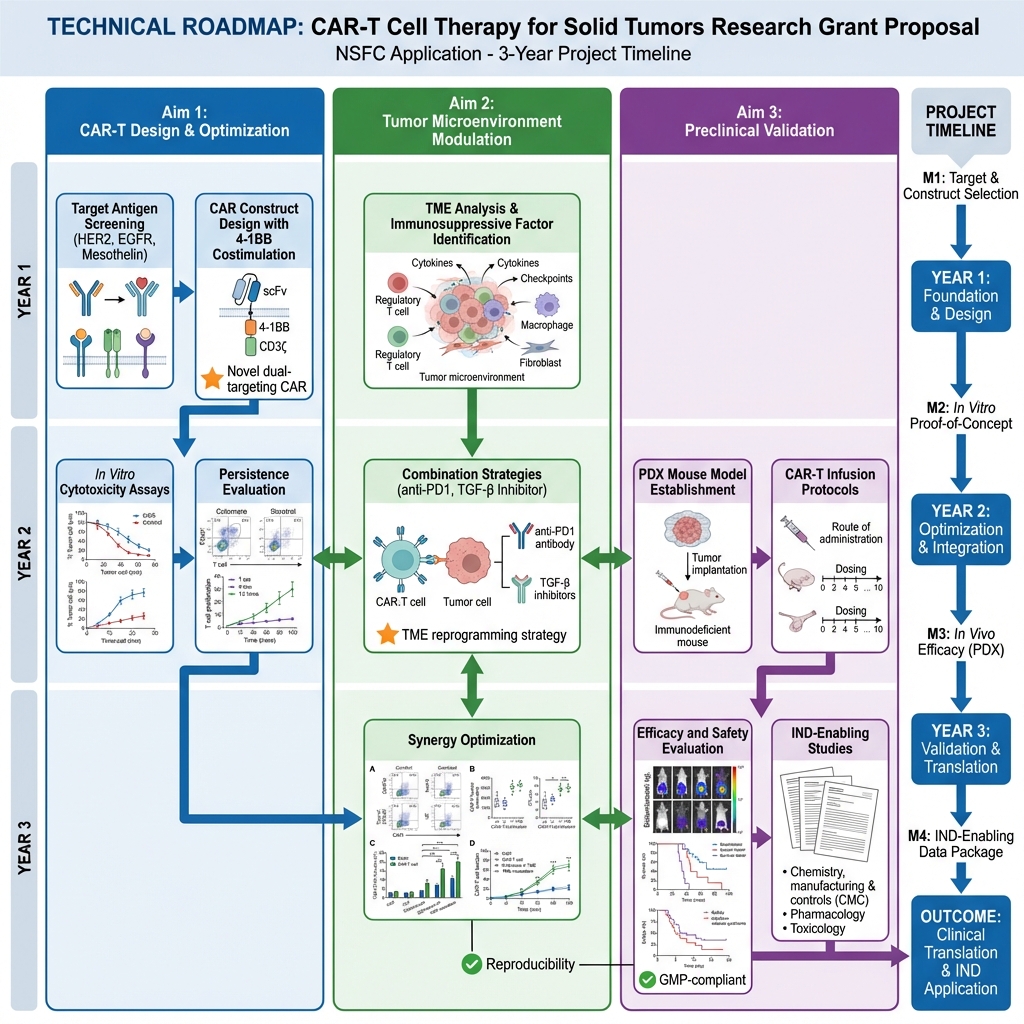}
  \caption{A technical roadmap for a multi-year research program on CAR-T cell therapy for solid tumors, generated using SciDraw. The figure integrates experimental aims, milestones, and project timelines in a structured visual format. Source: SciDraw~\citep{scidraw2025}.}
  \label{fig:roadmap}
\end{figure}

Collectively, these examples (Figures~\ref{fig:mechanism}--\ref{fig:roadmap}) demonstrate several strengths of domain-specific AI figure generation:
\begin{itemize}[nosep]
  \item Accurate and legible text rendering within complex diagrams;
  \item Consistent visual style across different figure types;
  \item Domain-appropriate use of color, iconography, and spatial organization;
  \item Sufficient detail for inclusion in academic manuscripts without post-editing.
\end{itemize}

\subsection{Selection Criteria for AI Figure Tools}

Based on our analysis, we recommend that researchers evaluate AI figure-generation tools using the following criteria:

\begin{enumerate}[nosep]
  \item \textbf{Output quality:} Does the tool produce figures that meet the visual standards of the target journal?
  \item \textbf{Domain specificity:} Does the tool understand the conventions of scientific illustration?
  \item \textbf{Text accuracy:} Can the tool reliably render labels, annotations, and legends?
  \item \textbf{Reproducibility support:} Does the tool archive generation parameters for disclosure?
  \item \textbf{Legal clarity:} What are the copyright and licensing terms for generated images?
  \item \textbf{Iterative workflow:} Can the researcher refine the output through conversation rather than starting from scratch?
\end{enumerate}

\section{Proposed Best-Practice Guidelines}
\label{sec:guidelines}

Based on our survey of publisher policies (Section~\ref{sec:policies}), analysis of community concerns (Section~\ref{sec:concerns}), and comparison of available tools (Section~\ref{sec:tools}), we propose the following best-practice guidelines for researchers using AI-generated figures in academic publications.

\subsection{Disclosure Statements}

We recommend a standardized three-part disclosure framework:

\begin{enumerate}
  \item \textbf{Methods Section:} Include a dedicated subsection titled ``\emph{AI-Assisted Figure Generation}'' that specifies:
  \begin{itemize}[nosep]
    \item The AI tool(s) used (name, version, and URL);
    \item Which specific figures were AI-generated or AI-assisted;
    \item A brief description of the generation process (e.g., text-to-image, iterative refinement);
    \item The date of generation and, if available, the model version.
  \end{itemize}

  \item \textbf{Figure Legends:} Each AI-generated figure should include a note in the legend, e.g., ``\emph{This figure was generated using [Tool Name] (version X.Y) with iterative prompt refinement. The generation prompt and metadata are available in the Supplementary Materials.}''

  \item \textbf{Cover Letter:} Briefly mention to the editor that AI tools were used for figure generation and that full disclosure is provided in the manuscript.
\end{enumerate}

\noindent\textbf{Example disclosure statement:}

\begin{quote}
\small
\emph{Figures~1, 3, and 5 in this manuscript were generated using SciDraw (\url{https://sci-draw.com}), an AI-powered scientific illustration platform. The figures were created through iterative prompt-based generation and refined through the platform's conversational interface. The full generation prompts and metadata are provided in Supplementary Table~S1. All figures were reviewed by the authors for scientific accuracy and visual fidelity.}
\end{quote}

\subsection{Human Review and Quality Control}

AI-generated figures should never be published without careful human review. We recommend the following quality-control checklist:

\begin{itemize}[nosep]
  \item \textbf{Scientific accuracy:} Do all labels, annotations, and spatial relationships correctly represent the intended science?
  \item \textbf{Visual consistency:} Is the figure consistent with the visual style of other figures in the manuscript?
  \item \textbf{Text legibility:} Are all text elements correctly spelled and legible at print resolution?
  \item \textbf{Data integrity:} If the figure contains any data-derived elements, have these been verified against the source data?
  \item \textbf{Bias check:} Does the figure inadvertently introduce visual biases (e.g., misleading proportions, suggestive color coding)?
\end{itemize}

\subsection{Aligning with Journal Policies}

We recommend the following workflow for ensuring compliance with journal-specific policies:

\begin{enumerate}[nosep]
  \item \textbf{Pre-submission:} Review the target journal's AI policy before figure generation. Identify any categorical restrictions (e.g., Cell Press's prohibition on AI-generated data figures).
  \item \textbf{During generation:} Use tools that support metadata archiving, such as SciDraw, to facilitate subsequent disclosure.
  \item \textbf{Pre-submission review:} Have a co-author independently verify all AI-generated figures for accuracy.
  \item \textbf{Submission:} Include disclosure in the methods section, figure legends, and cover letter as described above.
  \item \textbf{Post-acceptance:} Ensure that any journal-specific formatting requirements (e.g., resolution, file format) are met. Note that AI-generated figures may require format conversion.
\end{enumerate}

\subsection{Record Keeping}

For long-term reproducibility, we recommend that researchers:

\begin{itemize}[nosep]
  \item Archive the full prompt text and generation parameters for each AI-generated figure;
  \item Record the model name, version, and platform (e.g., ``SciDraw, Gemini 2.5 Flash backend, March 2026'');
  \item Store original output files at maximum resolution;
  \item Consider depositing generation metadata in a supplementary repository (e.g., Zenodo, Figshare).
\end{itemize}

\section{Conclusion}
\label{sec:conclusion}

AI-generated figures represent both an opportunity and a challenge for academic publishing. Our survey reveals that the policy landscape is rapidly evolving but remains fragmented, with significant variation in how publishers approach AI-generated visual content. We identify reproducibility, authorship, misinformation, and copyright as the primary concerns driving current policies.

The practical examples presented in this paper---generated using SciDraw~\citep{scidraw2025}, a platform specifically designed for academic scientific illustration---demonstrate that current AI tools are capable of producing figures that meet the visual and informational standards of academic publishing, particularly for schematic diagrams, research frameworks, and conceptual illustrations.

We propose a set of best-practice guidelines centered on transparent disclosure, rigorous human review, and systematic record keeping. We believe that widespread adoption of these guidelines can help the academic community realize the benefits of AI-assisted figure generation while maintaining scientific integrity.

Looking ahead, we call upon publishers to:
\begin{itemize}[nosep]
  \item Develop more specific and harmonized guidelines for AI-generated figures;
  \item Distinguish clearly between AI-generated data figures and AI-generated schematic/conceptual figures;
  \item Invest in tools and processes for detecting AI-generated content during peer review;
  \item Engage with the research community to develop evolving best practices.
\end{itemize}

The future of scientific illustration is likely to be hybrid, combining human expertise with AI capabilities. By establishing clear norms now, the academic community can ensure that this transition enhances, rather than undermines, the reliability and clarity of scientific communication.

\section*{Acknowledgments}

The author thanks the SciDraw development team for providing access to the platform and generating the example figures used in this paper. The figures in Figures~\ref{fig:mechanism}--\ref{fig:roadmap} were generated using SciDraw (\url{https://sci-draw.com}) for illustrative purposes.

\bibliographystyle{plainnat}

\begin{thebibliography}{20}

\bibitem[Bik et~al.(2016)]{bik2016prevalence}
Bik, E.~M., Casadevall, A., and Fang, F.~C. (2016).
\newblock The prevalence of inappropriate image duplication in biomedical
  research publications.
\newblock \emph{mBio}, 7(3):e00809--16.

\bibitem[Bindslev(2008)]{bindslev2008scientific}
Bindslev, H. (2008).
\newblock \emph{Scientific Visualization: The Visual Extraction of Knowledge
  from Data}.
\newblock Springer.

\bibitem[Cell Press(2024)]{cellpress2024ai}
Cell Press (2024).
\newblock Cell Press policy on AI-generated content in manuscripts.
\newblock \url{https://www.cell.com/cell/authors}, Accessed January 2026.

\bibitem[Elsevier(2024)]{elsevier2024ai}
Elsevier (2024).
\newblock The use of AI and AI-assisted technologies in writing for Elsevier.
\newblock \url{https://www.elsevier.com/about/policies-and-standards/the-use-of-generative-ai-and-ai-assisted-technologies-in-writing-for-elsevier}, Accessed January 2026.

\bibitem[Ioannidis(2005)]{ioannidis2005most}
Ioannidis, J.~P.~A. (2005).
\newblock Why most published research findings are false.
\newblock \emph{PLoS Medicine}, 2(8):e124.

\bibitem[Nature(2024)]{nature2024ai}
Nature (2024).
\newblock Artificial intelligence (AI) policy.
\newblock \url{https://www.nature.com/nature/editorial-policies}, Accessed January 2026.

\bibitem[PLOS(2024)]{plos2024ai}
PLOS (2024).
\newblock PLOS policy on the use of generative AI in published research.
\newblock \url{https://plos.org/resource/how-to-use-generative-ai-tools-responsibly/}, Accessed January 2026.

\bibitem[Ramesh et~al.(2022)]{ramesh2022hierarchical}
Ramesh, A., Dhariwal, P., Nichol, A., Chu, C., and Chen, M. (2022).
\newblock Hierarchical text-conditional image generation with CLIP latents.
\newblock \emph{arXiv preprint arXiv:2204.06125}.

\bibitem[Rombach et~al.(2022)]{rombach2022high}
Rombach, R., Blattmann, A., Lorenz, D., Esser, P., and Ommer, B. (2022).
\newblock High-resolution image synthesis with latent diffusion models.
\newblock In \emph{Proceedings of the IEEE/CVF Conference on Computer Vision
  and Pattern Recognition}, pages 10684--10695.

\bibitem[Rougier et~al.(2014)]{rougier2014ten}
Rougier, N.~P., Droettboom, M., and Bourne, P.~E. (2014).
\newblock Ten simple rules for better figures.
\newblock \emph{PLoS Computational Biology}, 10(9):e1003833.

\bibitem[Samuelson(2023)]{samuelson2023generative}
Samuelson, P. (2023).
\newblock Generative AI meets copyright.
\newblock \emph{Science}, 381(6654):158--161.

\bibitem[SciDraw(2025)]{scidraw2025}
SciDraw (2025).
\newblock SciDraw: AI-powered scientific illustration platform.
\newblock \url{https://sci-draw.com}.

\bibitem[Team et~al.(2023)]{team2023gemini}
Team, G., Anil, R., Borgeaud, S., Wu, Y., et~al. (2023).
\newblock Gemini: A family of highly capable multimodal models.
\newblock \emph{arXiv preprint arXiv:2312.11805}.

\bibitem[Thorp(2023)]{thorp2023chatgpt}
Thorp, H.~H. (2023).
\newblock ChatGPT is fun, but not an author.
\newblock \emph{Science}, 379(6630):313.

\bibitem[Tufte(2001)]{tufte2001visual}
Tufte, E.~R. (2001).
\newblock \emph{The Visual Display of Quantitative Information}.
\newblock Graphics Press, 2nd edition.

\bibitem[van Dis et~al.(2023)]{van2023chatgpt}
van Dis, E.~A.~M., Bollen, J., Zuidema, W., van Rooij, R., and Bockting, C.~L. (2023).
\newblock ChatGPT: Five priorities for research.
\newblock \emph{Nature}, 614(7947):224--226.

\end{thebibliography}

\end{document}